\documentclass[11pt,a4paper]{IEEEtran}
\usepackage{epsf}
\usepackage{amsmath, graphicx}
\usepackage{multirow,multicol}
\usepackage{caption}
\usepackage{float}
\usepackage{latexsym}
\usepackage{amssymb,amsfonts}
\usepackage{url}
\usepackage{color,soul}
\usepackage{subfigure}
\begin{document}

\title{Multimodal Sentiment Analysis: Addressing Key Issues and Setting up the Baselines}

\author{Soujanya Poria, Navonil Majumder, Devamanyu Hazarika, \\Erik Cambria, Alexander Gelbukh and Amir Hussain 
 \thanks{N.~Majumder and~A.~Gelbukh are with the CIC, Instituto Polit\'ecnico Nacional, Mexico City, Mexico.}% <-this % stops a space
        \thanks{S.~Poria and E.~Cambria is with the SCSE, Nanyang Technological Institute, Singapore.}% <-this % stops a space
        \thanks{D.~Hazarika is with the SOC, National University of Singapore, Singapore.}% <-this % stops a space
        \thanks{A.~Hussain is with the Edinburgh Napier University, UK.
        }% <-this % stops a space}
}

\maketitle

\begin{abstract}
    % 150-250 words (currently 101)

We compile baselines, along with dataset split, for multimodal sentiment
analysis. In this paper, we explore three different deep-learning based
architectures for multimodal sentiment classification, each improving upon the
previous. Further, we evaluate these architectures with multiple datasets with
fixed train/test partition. We also discuss some major issues, frequently
ignored in multimodal sentiment analysis research, e.g., role of
speaker-exclusive models, importance of different modalities, and
generalizability. This framework illustrates the different facets of analysis to
be considered while performing multimodal sentiment analysis and, hence, serves
as a new benchmark for future research in this emerging field.
\end{abstract}

\section{Introduction}

Emotion recognition and sentiment analysis is opening up numerous opportunities
pertaining social media in terms of understanding users preferences, habits, and
their contents \cite{porrev}. With the advancement of communication technology,
abundance of mobile devices, and the rapid rise of social media, a large amount
of data is being uploaded as video, rather than text \cite{cambig}. For example,
consumers tend to record their opinions on products using a webcam and upload
them on social media platforms, such as YouTube and Facebook, to inform the
subscribers of their views. Such videos often contain comparisons of products
from competing brands, pros and cons of product specifications, and other
information that can aid prospective buyers to make informed decisions.

The primary advantage of analyzing videos over mere text analysis, for detecting
emotions and sentiment, is the surplus of behavioral cues. Videos provide
multimodal data in terms of vocal and visual modalities. The vocal modulations
and facial expressions in the visual data, along with text data, provide
important cues to better identify true affective states of the opinion
holder. Thus, a combination of text and video data helps to create a better
emotion and sentiment analysis model.

Recently, a number of approaches to multimodal sentiment analysis producing
interesting results have been
proposed~\cite{poria2015deep,porcon}. However,
there are major issues that remain mostly unaddressed in this field, such as the
consideration of context in classification, effect of speaker-inclusive and
speaker-exclusive scenario, the impact of each modality across datasets, and
generalization ability of a multimodal sentiment classifier. Not tackling these
issues has presented difficulties in effective comparison of different
multimodal sentiment analysis methods. In this paper, we outline some methods
that address these issues and setup a baseline based on state-of-the-art
methods. We use a deep convolutional neural network (CNN) to extract features
from visual and text modalities.

This paper is organized as follows: Section~\ref{sec:related} provides a brief
literature review on multimodal sentiment analysis; Section~\ref{sec:feature-extraction}
briefly discusses the baseline methods; experimental results and discussion are
given in Section~\ref{exp};
finally, Section~\ref{sec:conclusion} concludes the paper.

\section{Related Work}
\label{sec:related}
In 1970, Ekman et al.~\cite{ekman1974universal} carried out extensive studies on
facial expressions. Their research showed that universal facial expressions are
able to provide sufficient clues to detect emotions. Recent studies on
speech-based emotion analysis~\cite{datcu2008semantic} have focused on
identifying relevant acoustic features, such as fundamental frequency (pitch),
intensity of utterance, bandwidth, and duration.

As to fusing audio and visual modalities for emotion recognition, two of the
early works were done by De Silva et al.~\cite{de1997facial} and Chen et
al.~\cite{chen1998multimodal}. Both works showed that a bimodal system yielded a
higher accuracy than any unimodal system. %More recent research on %audio-visual
%fusion for emotion recognition has been conducted at either feature
%level~\cite{kessous2010multimodal} or decision
%level~\cite{schuller2011recognizing}.

While there are many research papers on audio-visual fusion for emotion
recognition, only a few research works have been devoted to multimodal emotion
or sentiment analysis using text clues along with visual and audio
modalities. Wollmer et al.~\cite{wollmer2013YouTube} fused information from audio, visual and text
modalities to extract emotion and sentiment. Metallinou et
al.~\cite{metallinou2008audio} fused audio
and text modalities for emotion recognition. Both approaches relied on
feature-level fusion.

In this paper, we study the behavior of the method proposed in \cite{poria-EtAl:2017:Long} in the aspects rarely
addressed by other authors, such as speaker independence, generalizability of
the models and performance of individual modalities.

\section{Unimodal Feature Extraction}
\label{sec:feature-extraction}

% For the sake of a fair comparison with the state-of-the-art method CMN~\cite{hazarika-EtAl:2018:N18-1},
% we use identical unimodal features as \citet{hazarika-EtAl:2018:N18-1}.
For the unimodal feature extraction,
we follow the procedures by bc-LSTM~\cite{poria-EtAl:2017:Long}.

\subsection{Textual Feature Extraction}
\label{sec:text-feat-extr}

We employ convolutional neural networks (CNN) for textual feature extraction. Following
\cite{kim2014convolutional}, we obtain n-gram features from each utterance using three distinct
convolution filters
of sizes 3, 4, and 5 respectively, each having 50 feature-maps. Outputs are then subjected to max-pooling followed by rectified linear unit (ReLU) activation. These activations are concatenated and fed to a $100$ dimensional dense layer, which is regarded as the textual utterance representation. This network is trained at utterance level with the emotion labels.

\subsection{Audio and Visual Feature Extraction}
\label{sec:visu-feat-extr}
Identical to \cite{poria-EtAl:2017:Long}, we use 3D-CNN and openSMILE~\cite{eyben2010opensmile} for visual and acoustic feature extraction, respectively.

\subsection{Fusion}
\label{sec:fusion}
In order to fuse the information extracted from different modalities, we
concatenated the feature vectors representative of the given modalities and sent
the combined vector to a classifier for the classification. This scheme of fusion is
called feature-level fusion. Since, the fusion involved concatenation and no
overlapping, merge, or combination, scaling and normalization of the features
were avoided. We discuss the results of this fusion in Section~\ref{exp}.
\subsection{Baseline Method}
\subsubsection{bc-LSTM}
We follow the method bc-LSTM~\cite{poria-EtAl:2017:Long} where they used a biredectional LSTM to capture the context from
the surrounding utterances to generate context-aware utterance
representation.
%\begin{figure}[h]
%    \centering
%    \includegraphics[width=\linewidth]{architecture.png}
%    \caption{Overall architecture of the proposed method.}
%    \label{fig:arch}
%\end{figure}
\subsubsection{SVM}
After extracting the features, we merged and sent to a SVM with RBF kernel for the final classification.
\section{Experiments and Observations}
\label{exp}

In this section, we discuss the datasets and the experimental settings. Also, we
analyze the results yielded by the aforementioned methods.

\subsection{Datasets}

\subsubsection{Multimodal Sentiment Analysis Datasets}

For our experiments, we used the MOUD dataset, developed by Perez-Rosas et
al.~\cite{perez2013utterance}. They collected 80 product review and
recommendation videos from YouTube. Each video was segmented into its utterances
(498 in total) and each of these was categorized by a sentiment label (positive,
negative and neutral). On average, each video has 6 utterances and each
utterance is 5 seconds long. In our experiment, we did not consider neutral
labels, which led to the final dataset consisting of 448 utterances. We dropped
the neutral label to maintain consistency with previous work. In a similar
fashion, Zadeh et al.~\cite{zadeh2016multimodal} constructed a multimodal
sentiment analysis dataset called multimodal opinion-level sentiment intensity
(MOSI), which is bigger than MOUD, consisting of 2199 opinionated utterances, 93
videos by 89 speakers. The videos address a large array of topics, such as
movies, books, and products. In the experiment to address the generalizability
issues, we trained a model on MOSI and tested on MOUD. Table~\ref{tab:dataset}
shows the split of train/test of these datasets.

\subsubsection{Multimodal Emotion Recognition Dataset}

The IEMOCAP database~\cite{busso2008iemocap} was collected for the purpose of
studying multimodal expressive dyadic interactions. This dataset contains 12
hours of video data split into 5 minutes of dyadic interaction between
professional male and female actors. Each interaction session was split into
spoken utterances. At least 3 annotators assigned to each utterance one emotion
category: \emph{happy, sad, neutral, angry, surprised, excited, frustration,
disgust, fear} and \emph{other}. In this work, we considered only the utterances
with majority agreement (i.e., at least two out of three annotators labeled the
same emotion) in the emotion classes of \emph{angry}, \emph{happy}, \emph{sad},
and \emph{neutral}. Table~\ref{tab:dataset} shows the split of train/test of this
dataset.

\begin{table}[h]
\small
	\begin{center}
		\begin{tabular}{|*{6}{c|}}
			\hline
			\multicolumn{2}{|c|}{\multirow{2}{*}{Dataset}} & \multicolumn{2}{c|}{Train} & \multicolumn{2}{c|}{Test}\\ \cline{3-6}
			\multicolumn{2}{|c|}{}& \emph{utterance}&\emph{video}&\emph{utterance}&\emph{video}\\ \hline
			\multicolumn{2}{|c|}{IEMOCAP}& 4290& 120& 1208& 31\\ \hline
			\multicolumn{2}{|c|}{MOSI}& 1447& 62& 752& 31\\ \hline
			\multicolumn{2}{|c|}{MOUD}& 322& 59& 115& 20\\ \hline
			\multicolumn{2}{|c|}{MOSI $\rightarrow$ MOUD}&2199&93&437&79\\ \hline
		\end{tabular}
	\end{center}
	\caption { Person-Independent Train/Test split details of each dataset ($\approx$ 70/30 \% split). Note: X$\rightarrow$Y represents train: X and test: Y; Validation sets are extracted from the shuffled train sets using 80/20 \% train/val ratio. }
	\label{tab:dataset}
\end{table}

\subsection{Speaker-Exclusive Experiment}

Most of the research on multimodal sentiment analysis is performed with datasets
having common speaker(s) between train and test splits. However, given this
overlap, results do not scale to true generalization.
%because each individual is unique in his/her own way of expressing emotions and
% sentiments, finding generic, person-independent features for sentimental
% analysis is very important.
In real-world applications, the model should be robust to speaker
variance. Thus, we performed speaker-exclusive experiments to emulate unseen
conditions. This time, our train/test splits of the datasets were completely
disjoint with respect to speakers. While testing, our models had to classify
emotions and sentiments from utterances by speakers they have never seen
before. Below, we elaborate this speaker-exclusive experiment:

\begin{itemize}

\item \textbf{IEMOCAP:} As this dataset contains 10 speakers, we performed a
10-fold speaker-exclusive test, where in each round exactly one of the speakers
was included in the test set and missing from train set. The same SVM model was
used as before and accuracy was used as performance metric.

\item \textbf{MOUD:} This dataset contains videos of about 80 people reviewing
various products in Spanish. Each utterance in the video has been labeled as
\emph{positive, negative}, or \emph{neutral}. In our experiments, we consider
only samples with \emph{positive} and \emph{negative} sentiment labels. The
speakers were partitioned into 5 groups and a 5-fold person-exclusive experiment
was performed, where in every fold one out of the five group was in the test
set. Finally, we took average of the accuracy to summarize the results
(Table~\ref{tab:speakerindep}).

\item \textbf{MOSI:} MOSI dataset is rich in sentimental expressions, where 93
people review various products in English. The videos are segmented into clips,
where each clip is assigned a sentiment score between $-3$ to $+3$ by five
annotators. We took the average of these labels as the sentiment polarity and
naturally considered two classes (\emph{positive} and \emph{negative}). Like
MOUD, speakers were divided into five groups and a 5-fold person-exclusive
experiment was run. For each fold, on average 75 people were in the training set
and the remaining in the test set. The training set was further partitioned and
shuffled into 80\%--20\% split to generate train and validation sets
for parameter tuning.

\end{itemize}

\subsubsection{Speaker-Inclusive vs. Speaker-Exclusive}

\begin{figure}[h]
    \centering
    \includegraphics[width=0.45\linewidth]{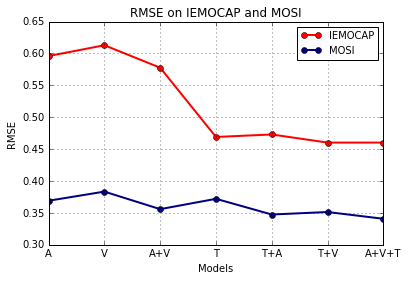}%
    \includegraphics[width=0.45\linewidth]{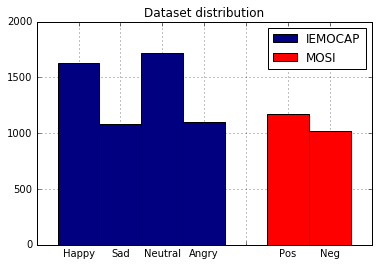}
    \includegraphics[width=0.45\linewidth]{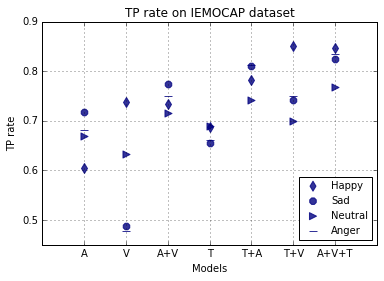}
    \includegraphics[width=0.45\linewidth]{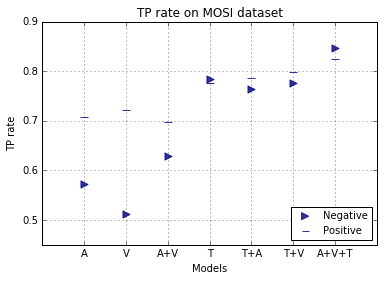}
    
    \caption{Experiments on IEMOCAP and MOSI datasets. The top-left figure shows the RMSE of the models on IEMOCAP and MOSI. The top-right figure shows the dataset distribution. Bottom-left and bottom-right figures present TP-rate on of the models on IEMOCAP and MOSI dataset, respectively. }
    \label{fig:fig1}
\end{figure}

In comparison with the speaker-inclusive experiment, the speaker-exclusive
setting yielded inferior results. This is caused by the absence of knowledge
about the speakers during the testing phase. Table~\ref{tab:speakerindep} shows
the performance obtained in the speaker-inclusive experiment. It can be seen
that audio modality consistently performs better than visual modality in both
MOSI and IEMOCAP datasets. The text modality plays the most important role in
both emotion recognition and sentiment analysis. The fusion of the modalities
shows more impact for emotion recognition than for sentiment analysis. Root mean
square error (RMSE) and TP-rate of the experiments using different modalities on
IEMOCAP and MOSI datasets are shown in Fig.~\ref{fig:fig1}.

\begin{table*}[h]
    \centering
    \begin{tabular}{|c|c|c|c|c|c|c|}
      \hline
      \multirow{2}{*}{
      \begin{tabular}{c}
        Modality\\ Combination
      \end{tabular}
}
      & \multicolumn{2}{c|}{IEMOCAP} & \multicolumn{2}{c|}{MOUD} & \multicolumn{2}{c|}{MOSI} \\
      \cline{2-7} & Sp-In & Sp-Ex & Sp-In & Sp-Ex & Sp-In & Sp-Ex \\
      \hline
      A         & 66.20         & 51.52          & -- & 53.70          & 64.00       & 57.14 \\
      V         & 60.30         & 41.79          & -- & 47.68          & 62.11       & 58.46 \\
      T         & 67.90         & 65.13          & -- & 48.40          & 78.00       & 75.16 \\
      \hline
      T + A     & 78.20         & 70.79          & -- & 57.10          & 76.60       & 75.72 \\
      T + V     & 76.30         & 68.55          & -- & 49.22          & 78.80       & 75.06 \\
      A + V     & 73.90         & 52.15          & -- & 62.88          & 66.65       & 62.4 \\
      \hline
      T + A + V & \textbf{81.70}& \textbf{71.59} & -- & \textbf{67.90} & \textbf{78.80}& \textbf{76.66} \\
      \hline
    \end{tabular}
    \caption{Accuracy reported for speaker-exclusive (Sp-Ex) and
speaker-inclusive (Sp-In) split for Concatenation-Based Fusion. \emph{IEMOCAP:}
10-fold speaker-exclusive average. \emph{MOUD:} 5-fold speaker-exclusive
average. \emph{MOSI:} 5-fold speaker-exclusive average. \emph{Legend:} A stands
for Audio, V for Video, T for Text.}
    \label{tab:speakerindep}
\end{table*}

\subsection{Contributions of the Modalities}

\begin{figure}[t]
    \centering
    \includegraphics[width=\linewidth]{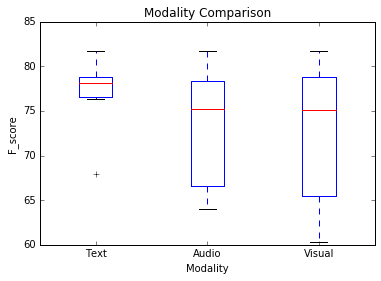}%
    \caption{Performance of the modalities on the datasets. Red line indicates
the median of the accuracy.}
    \label{fig:fig2}
\end{figure}

As expected, bimodal, and trimodal models have performed better than unimodal
models in all experiments. Overall, audio modality has performed better than
visual on all datasets. Except for MOUD dataset, the unimodal performance of
text modality is substantially better than other two modalities
(Fig.~\ref{fig:fig2}).

\subsection{Generalizability of the Models}

To test the generalization ability of the models, we trained the framework on
MOSI dataset in speaker-exclusive fashion and tested with MOUD dataset. From
Table~\ref{tab:gen}, we can see that the trained model with MOSI dataset
performed poorly with MOUD dataset.

This is mainly due to the fact that reviews in MOUD dataset had been recorded in
Spanish, so both audio and text modalities miserably fail in recognition, as
MOSI dataset contains reviews in English. A more comprehensive study would be to
perform generalizability tests on datasets of the same language. However, we
were unable to do this for the lack of benchmark datasets. Also, similar
experiments of cross-dataset generalization was not performed on emotion
detection, given the availability of only a single dataset (IEMOCAP).

\begin{table}[t]
    \centering
    \begin{tabular}{c|c|c}
      \hline
      \multirow{2}{*}{Modality Combination}    & \multicolumn{2}{c}{Accuracy} \\
      \cline{2-3} & SVM & bc-LSTM \\
      \hline
      T         & 46.5\%  & \bf46.9\% \\
      V         & 43.3\%  & \bf49.6\% \\
      A         & 42.9\%  & \bf47.2\% \\
      T + A     & 50.4\%  & \bf51.3\% \\
      T + V     & 49.8\%  & \bf49.8\% \\
      A + V     & 46.0\%  & \bf49.6\% \\
      T + A + V & 51.1\%  & \bf52.7\% \\
      \hline
    \end{tabular}
    \caption{\textbf{Cross-dataset results: }Model (with previous
      configurations) trained on MOSI dataset and tested on MOUD dataset.}
    \label{tab:gen}
\end{table}

\subsection{Comparison among the Baseline Methods}
\label{sec:comp-among-meth}

Table~\ref{tab:consolidated-results} consolidates and compares performance of
all the baseline methods for all the datasets. We evaluated SVM and bc-LSTM fusion with MOSI, MOUD, and IEMOCAP dataset.

From Table~\ref{tab:consolidated-results}, it is clear that bc-LSTM
performs better than SVM across all the experiments. So, it is
very apparent that consideration of context in the classification process has
substantially boosted the performance.

\begin{table*}[h]
    \centering
    \begin{tabular}{|c||c|c||c|c||c|c|}
      \hline
      \multirow{2}{*}{
      \begin{tabular}{c}
        Modality\\ Combination
      \end{tabular}
}
      & \multicolumn{2}{c||}{IEMOCAP} & \multicolumn{2}{c||}{MOUD} & \multicolumn{2}{c|}{MOSI} \\
      \cline{2-7} & SVM & bc-LSTM & SVM & bc-LSTM & SVM & bc-LSTM\\
      \hline
      A         & 52.9 & \bf57.1 & 51.5 & \bf59.9 & 58.5 & \bf60.3  \\
      V         & 47.0 & \bf53.2 & 46.3 & \bf48.5 & 53.1 & \bf55.8 \\
      T         & 65.5 & \bf73.6 & 49.5 & \bf52.1 & 75.5 & \bf78.1\\
      T + A     & 70.1 & \bf75.4 & 53.1 & \bf60.4 & 75.8 & \bf80.2\\
      T + V     & 68.5 & \bf75.6 & 50.2 & \bf52.2 & 76.7 & 79.3\\
      A + V     & 67.6 & \bf68.9 & 62.8 & \bf65.3 & 58.6 & 62.1\\
      T + A + V & 72.5 & \bf76.1 & 66.1 & \bf68.1 & 77.9 & 80.3\\
      \hline
    \end{tabular}
    \caption{Accuracy reported for speaker-exclusive
classification. \emph{IEMOCAP:} 10-fold speaker-exclusive average. \emph{MOUD:}
5-fold speaker-exclusive average. \emph{MOSI:} 5-fold speaker-exclusive
average. \emph{Legend:} A represents Audio, V represents Video, T represents
Text.}
    \label{tab:consolidated-results}
\end{table*}

\subsection{Visualization of the Datasets}

\begin{figure}[h]
    \centering
    \begin{subfigure}
        \centering
        \includegraphics[width=0.5\textwidth]{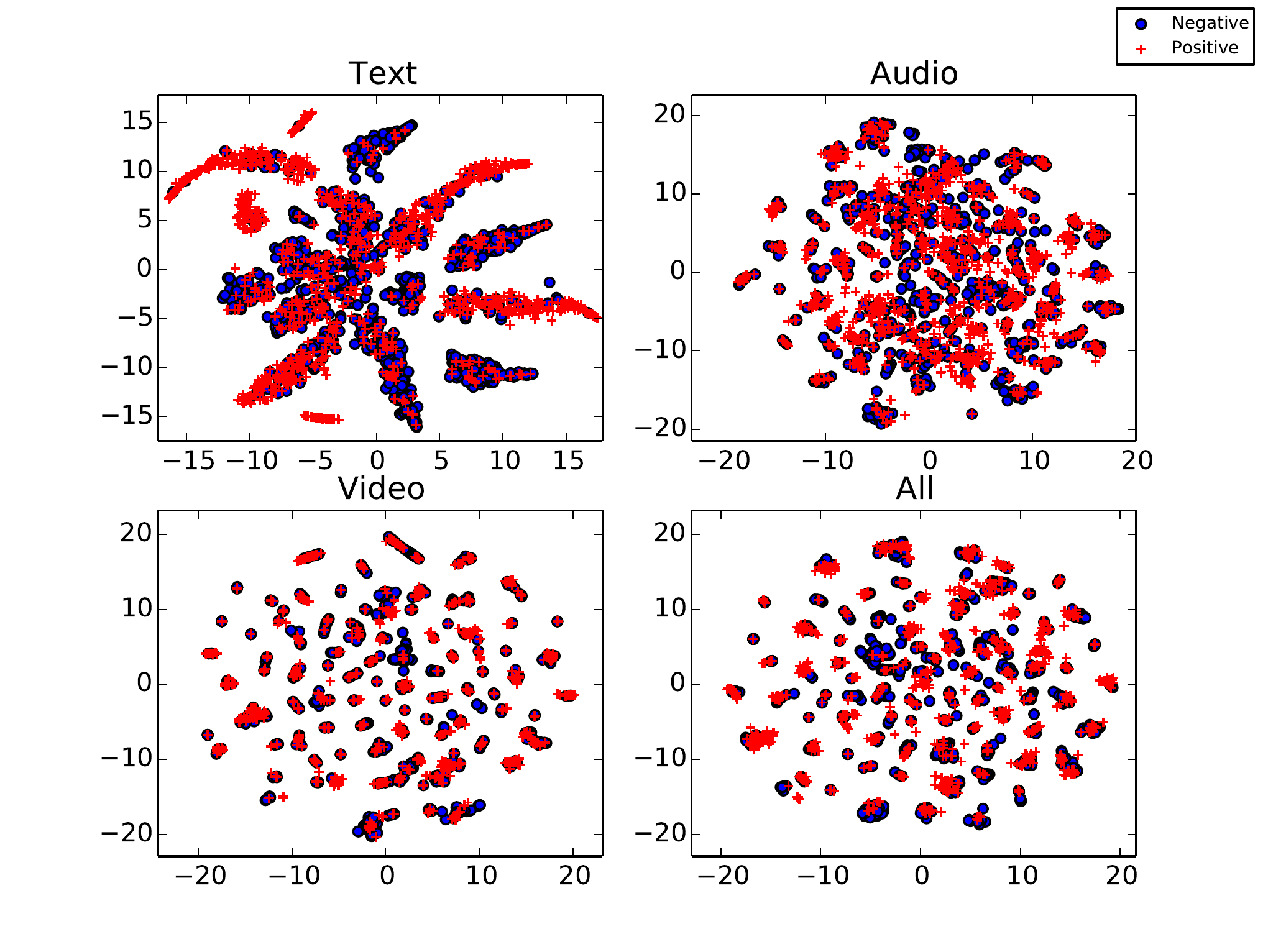}%
    \end{subfigure}
    \begin{subfigure}
        \centering
        \includegraphics[width=0.5\textwidth]{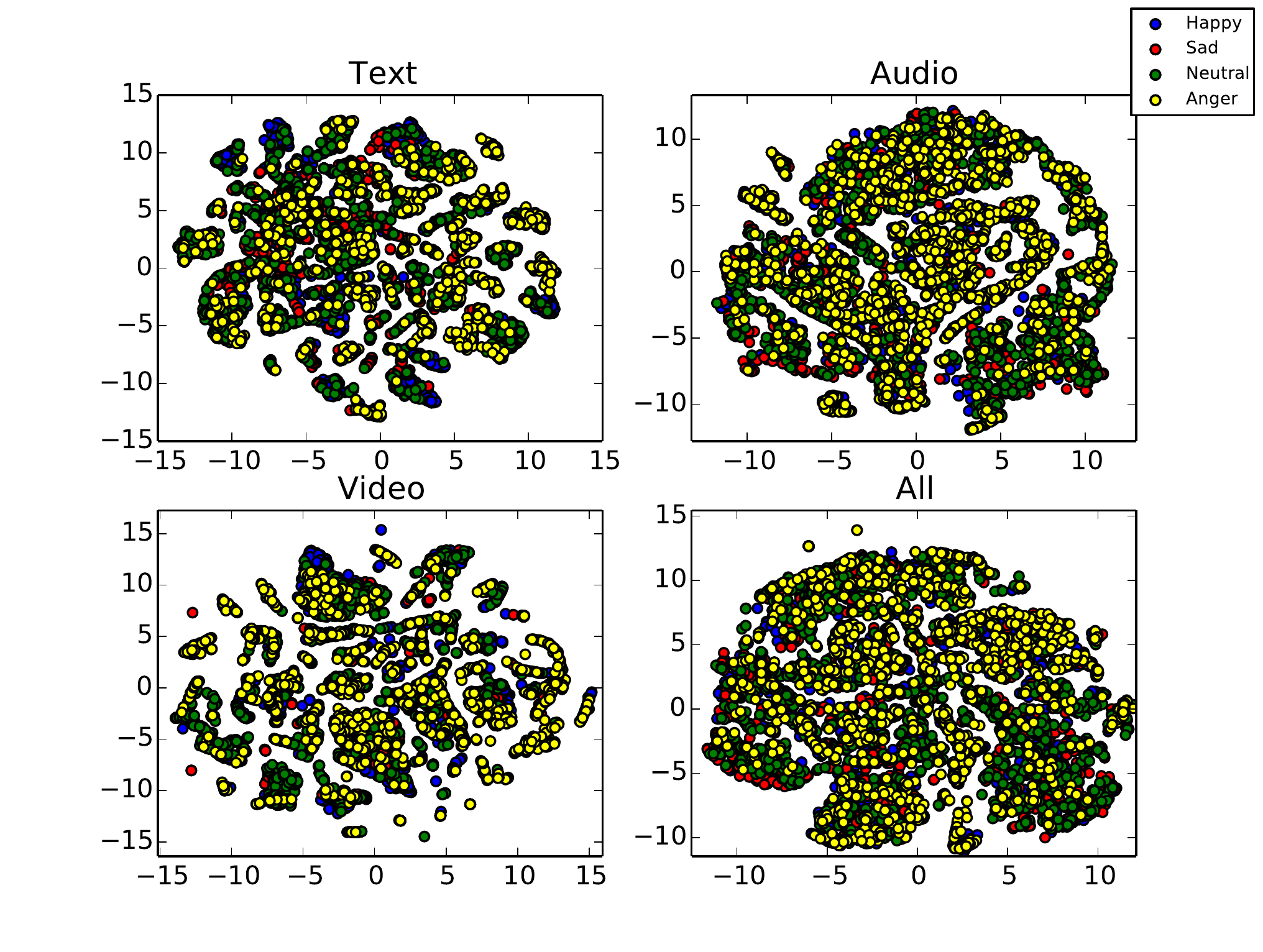}%
    \end{subfigure}
    \caption{T-SNE 2D visualization of MOSI and IEMOCAP datasets when unimodal
      features and multimodal features are used.}
    \label{fig:fig3}
\end{figure}

MOSI visualizations present information regarding dataset distribution within
single and multiple modalities (Fig.~\ref{fig:fig3}). For the textual and audio
modalities, comprehensive clustering can be seen with substantial
overlap. However, this problem is reduced in the video and all modalities
scenario with structured de\-clustering but overlap is reduced only in
multimodal. This forms an intuitive explanation of the improved performance in
the multi\-modal scenario. IEMOCAP visualizations provide insight for the
4-class distribution for uni and multimodal scenario, where clearly the
multi\-modal distribution has the least overlap (increase in red and blue
visuals, apart from the rest) with sparse distribution aiding the classification
process.

\section{Conclusion}
\label{sec:conclusion}

We have presented useful baselines for multimodal sentiment analysis and
multimodal emotion recognition. We also discussed some major aspects of
multimodal sentiment analysis problem, such as the performance in the
unknown-speaker setting and the cross-dataset performance of the models.

Our future work will focus on extracting semantics from the visual features,
relatedness of the cross-modal features and their fusion. We will also include
contextual dependency learning in our model to overcome the limitations
mentioned in the previous section.

\bibliographystyle{abbrv}
\bibliography{benchmarking-multimodal-sentiment-analysis}

\end{document}